\documentclass[sigconf]{acmart}

 \usepackage{algorithm} 
\usepackage{algpseudocode}

\usepackage{tikz}
\usetikzlibrary{automata,positioning}
\usetikzlibrary{decorations.pathreplacing,decorations.markings,shapes.geometric}
\usetikzlibrary{shapes,arrows}
\usetikzlibrary{backgrounds,calc,positioning}

\usepackage{tabularx}       
\usepackage{float}          
\usepackage{booktabs}       

\usepackage{enumerate}

\def\be{ \begin{equation} }
\def\ee{ \end{equation} }
\def\bea{ \begin{eqnarray} }
\def\eea{ \end{eqnarray} }

\def\b0{{\bf 0}}

\catcode`,\active

\catcode`\,12

\theoremstyle{remark}

\setcopyright{none}

\acmDOI{}

\acmISBN{}

\acmPrice{}

\begin{document}
\title{Quantum Federated Learning Experiments in the Cloud with Data Encoding}
 \author{Shiva Raj Pokhrel,
Naman Yash,
Jonathan Kua,
Gang Li and
Lei Pan}

\renewcommand{\shortauthors}{Pokhrel et al.}

\begin{abstract}
Quantum Federated Learning (QFL) is an emerging concept that aims to unfold federated learning (FL) over quantum networks, enabling collaborative quantum model training along with local data privacy. We explore the challenges of deploying QFL on cloud platforms, emphasizing quantum intricacies and platform limitations. The proposed data-encoding-driven QFL, with a proof of concept (GitHub Open Source) using genomic data sets on quantum simulators, shows promising results.\footnote{Authors are from IoT \& SE Lab, School of IT, Deakin University, shiva.pokhrel@deakin.edu.au.} 
\end{abstract}

\maketitle

\section{Introduction} \label{S:Intro}

Quantum computing has unlocked unprecedented computational capabilities, offering solutions to problems beyond classical computers' reach~\cite{gill2022quantum}. This breakthrough holds particular promise in machine learning, where quantum algorithms can vastly accelerate data processing, impacting sectors like healthcare, finance, and cybersecurity~\cite{QFLRnC, QFLwQD}. By harnessing quantum computing, we can tackle previously insurmountable challenges, marking a significant milestone in computational science and its practical applications.

Our focus lies on Quantum Federated Learning (QFL), a frontier merging federated learning (FL) principles with quantum machine learning (QML) over quantum networks~\cite{caro2022generalization}. FL, designed to train models on decentralized devices while preserving data privacy, offers a potent solution for data analysis. By marrying FL with QML, our goal is to enhance computational efficiency and model performance within the quantum realm, all while safeguarding data privacy. Our key contribution lies in crafting and implementing a QFL algorithm, leveraging the potential of cloud-based quantum computing platforms. The analysis of various quantum cloud service providers is crucial to grasp the unique challenges and opportunities in this burgeoning field. Understanding existing quantum cloud infrastructures' capabilities, limitations, and progression roadmaps is paramount for translating FL principles into practical quantum implementations.

In classic FL, each client trains its model locally, sending model parameters to a global server for aggregation without sharing raw data~\cite{9079513}. Our paper demonstrates the feasibility of QFL using IBM's Qiskit quantum computing library, distinguishing itself by integrating FL into QML, specifically in cloud-based platforms. Our contributions encompass an in-depth analysis of current cloud-based quantum resources for QFL suitability and a proposal for a data-encoding-driven QFL implementation. By comparing state-of-the-art FL techniques, we aim to surmount challenges associated with deploying QFL algorithms.

\begin{figure}[t]
\centering
\includegraphics[width=0.9\columnwidth]{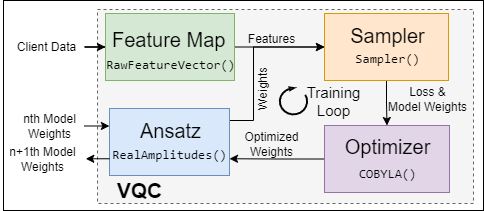}
\caption{\rm A high level view of local learning in the proposed QFL Process consisting of several key components. The \textit{Feature Map} ingests input data and encodes it into a quantum state. Following this, the \textit{Ansatz} comes into play as a parameterized quantum circuit, with its parameters being iteratively fed by the\textit{ Optimizer}--optimization objective function is driven by the outcomes from the \textit{Sampler}.}
\vspace{- 6 mm}
\label{fig:qfl_process}
\end{figure}

The novelty of this research lies in its potential to expedite QFL adoption, paving the way for quantum-enhanced machine learning models over the cloud trained efficiently in a distributed setting while upholding local privacy. We design a novel process, as illustrated with Qiskit components in Fig.~\ref{fig:qfl_process}, that can be perceived as approximately transforming input data into a quantum state, exploring and exploiting it using a customizable parameterized quantum circuit, and iteratively optimizing the parameters to steer and achieve the desired outcome based on the global objective function. 
The background details are discussed later in Sec.~\ref{Sec:3} with implementation details deferred later in Sec.~\ref{Sec:4}.

In the proposed QFL realization, clients transform their unique data into quantum states using a Feature Map, then process them with a parameterized quantum circuit (Ansatz) where local training is conducted using Qiskit and the updated weights are aggregated centrally, and global weights are returned to clients for local model updates.

\section{State of quantum computing}\label{Sec:2}

Quantum computing and QML promise dramatic speedups over classical machines for certain tasks. A prime example is the popular Shor's algorithm for factoring integers, where quantum computing provides an exponential advantage over the best-known classical methods \cite{Shor94}. Major tech firms are currently racing to build proprietary quantum hardware and offer cloud access. However, despite rapid progress quantum hardware development is still in nascent development stages \cite{QFLRnC}. 

Major tech companies like IBM, Amazon and Microsoft now offer cloud-based access to quantum computers. This allows users to experiment with the latest quantum algorithms without investing in specialized labs or hardware. IBM has emerged as an early front-runner, recently unveiling a 433-qubit quantum processor that currently reigns as the most powerful quantum system.\footnote{IBM unveils 400 qubit-plus quantum processor and next-generation IBM Quantum System Two, \href{https://newsroom.ibm.com/2022-11-09-IBM-Unveils-400-Qubit-Plus-Quantum-Processor-and-Next-Generation-IBM-Quantum-System-Two}{link}}  In contrast, Amazon and Microsoft provide access to a diverse range of third-party quantum processors, like IONQ's Harmony and Rigetti's Ankaa-1. By using AWS Braket and Azure Quantum, researchers can test algorithms on systems like IONQ's 32-qubit Harmony or Rigetti's 80-qubit Ankaa processor. As underlying quantum technologies continue rapidly improving, we should see more powerful processors integrated into the cloud ecosystems~\cite{AzurePricing, AmazonBraket}.

QML shows promise but remains in early developmental stages. While programming frameworks like IBM’s Qiskit are emerging as robust tools, there is a lack of standardization of different software libraries and limited developer forums and up-to-date resources which pose challenges to newcomers in the field. Cloud-based access aims to increase experimentation by removing on-premises infrastructure barriers. However, resource contention on shared quantum hardware can result in prolonged queuing delays before algorithms execute. This severely impacts techniques like QFL ( a full-fledge federation of QML) which require repeated access to quantum circuits for the model training.

While QFL holds immense potential, the current landscape faces challenges in hardware development, software standardization, and efficient cloud-based access for QFL. Looking at specific QML roadmaps, IBM aims to unveil Quantum System Two in 2024, a modular platform paving the way for a scalable quantum-centric ecosystem. By 2025, IBM targets Quantum Condor, a processor with over 1,000 qubits. IONQ focuses on practical quantum devices, projecting increased algorithmic qubits in their Forte and Tempo systems. Google's focus on increasing qubits, achieving a logical qubit prototype, and forming industry partnerships showcases its commitment to practical quantum applications. Meanwhile, Rigetti Computing's strides include launching Ankaa and Lyra processors and collaborations to enhance quantum cloud services, with plans to deploy testbeds at Fermilab by 2025.

Developed by IBM, Qiskit\footnote{https://qiskit-community.github.io/qiskit-machine-learning/} can play an important role in driving the substantial research and development progress of QFL. Its capabilities in quantum simulation and QML are key factors in overcoming QFL operationalization challenges and pushing the boundaries of this emerging field.


\section{Operationalizing QFL in Qiskit}\label{Sec:3}
In the realm of quantum computing, Qiskit's quantum simulators~\cite {IonQ_simulator_2023} emerge as indispensable tools, empowering developers to test quantum algorithms on classical computers. These simulators serve as flexible platforms that facilitate robust algorithm development and testing processes. Their accessibility benefits users of all levels of proficiency, reducing the iterative journey of algorithm refinement~\cite{kashifQiskitSim}. By providing a means to test algorithms on classical machines, they eliminate hardware access barriers while faithfully emulating the expected behavior of actual quantum systems. This capability accelerates rapid prototyping, empowering developers to iterate and refine without incurring costly overhead. Embracing Qiskit's quantum simulators is not just a choice, it is a strategic imperative for advancing distributed quantum algorithms development efficiently and effectively.


\subsection{Modeling QML Libraries}

Advances in the Qiskit's QML libraries will be cutting-edge contributions to advancing QFL. The existing library includes pre-built implementations of classifiers like the Variational Quantum Classifier (\href{https://qiskit-community.github.io/qiskit-machine-learning/stubs/qiskit_machine_learning.algorithms.VQC.html#}{VQC}) and the Quantum Support Vector Regressor (\href{https://qiskit-community.github.io/qiskit-machine-learning/stubs/qiskit_machine_learning.algorithms.QSVR.html#}{QSVR}). Extending these classifiers are vital in constructing QFL models, which form the foundation for success of the framework proposed in this paper. By offering ready-to-use modules, Qiskit simplifies the development cycle, allowing researchers and developers to focus on tailoring QFL models to specific application requirements.


Fig.~\ref{fig:qfl_process} illustrates the quantum machine learning process enabled by extending the \href{https://qiskit-community.github.io/qiskit-machine-learning/stubs/qiskit_machine_learning.algorithms.VQC.html#}{VQC}. In our QFL implementation, the VQC uses \texttt{RawFeatureVector()} as a feature map to encode classical data into quantum states. This feature map transforms classical input vectors into quantum states through a series of quantum gates, effectively preparing the quantum data for analysis. The \texttt{Sampler()} component executes the quantum circuits on various backends, which can be either simulators or real quantum hardware, thereby providing the flexibility to test and run quantum machine learning models in different environments. As shown in Fig.~\ref{fig:qfl_process}, the ansatz is a parameterized quantum circuit that is repeatedly run, where each iteration or'rep' applies a sequence of gates whose parameters are adjusted in each step of the optimization process. The \texttt{RealAmplitudes()} in ansatz is used within variational circuits to construct quantum states that are optimized for our QFL operationalization. The optimization is performed by employing the \texttt{COBYLA()} algorithm, a gradient-free optimizer suitable for the noisy conditions of current quantum technology. \texttt{COBYLA} iteratively adjusts the parameters of the ansatz in an attempt to minimize a loss function, guiding the model towards the most accurate classifications possible given the input data. This synergy of components represents the architecture of a VQC, which transforms classical data for quantum processing, leverages quantum computation for machine learning tasks, and optimizes its performance over multiple iterations.\footnote{https://qiskit-community.github.io/qiskit-machine-learning/}

 It is worth noting that Qiskit's quantum simulators and libraries collectively empower researchers and developers to develop and evaluate new QML and QFL algorithms easily. Furthermore, the inclusion of extended libraries within Qiskit shows a broader commitment to the evolution of distributed quantum machine learning algorithms.

\subsection{Implementation in Qiskit}\label{Sec:4}

In our pursuit of new strategies to operationalize QFL, we embark on an evolutionary journey from traditional FL. Witnessing the decentralization of model training in FL, where clients actively contribute while preserving data privacy, marks a pivotal phase. However, the transition to QFL introduces practical challenges demanding a fresh perspective.

Currently, achieving QFL with quantum devices on each client poses significant hurdles, primarily due to the absence of small-scale quantum devices on client-side platforms. The quantum computing landscape is dominated by technologies like superconducting qubits and ion trap-based quantum computers, which lack mobility.

In our current QFL implementation, client and global model training occurs primarily on the server. Clients provide data utilized for training on the server, which then returns client models. This orchestration, occurring after each epoch, represents a pragmatic compromise balancing quantum computing capabilities with the absence of client-side quantum devices and edge computing limitations.

\begin{figure}[t]
\centering
\includegraphics[width=0.8\columnwidth]{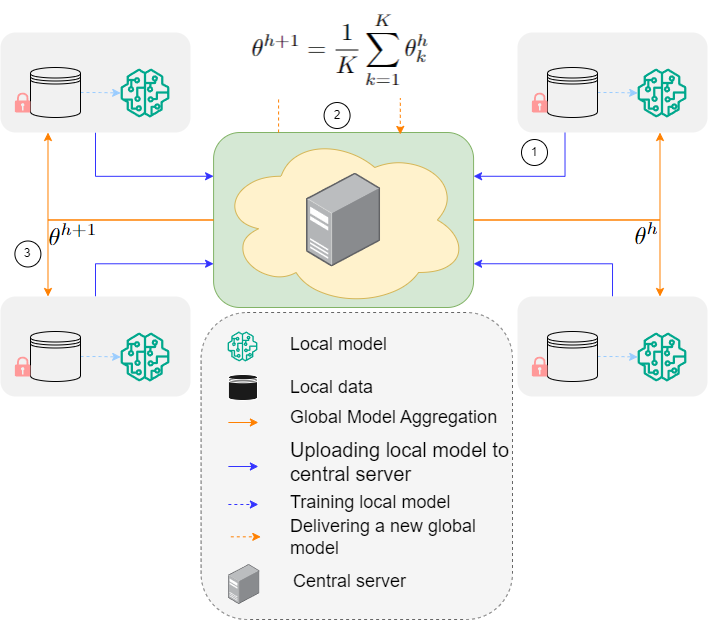}
\caption{\rm \textit{QFL Process.} Each client in the setup trains models locally and only shares its parameters with a central server (step 1). The server aggregates these to enhance the global model(step 2), then circulates the updated parameters back to clients(step 3), preserving data privacy as client data stays local.}
\label{fig:simple_fl_process}
\end{figure}

To pioneer the operationalization of QFL, we leverage IBM's quantum cloud platform, tapping into its array of quantum processors and simulators. Utilizing Qiskit's simulation capabilities forms the foundation for implementing the QFL framework, opting for established frameworks such as QFedAvg. Our proposed design and implementation of QFL is shown in Figure \ref{fig:simple_fl_process} and is detailed in the following.

We consider a server, as shown in Figure \ref{fig:simple_fl_process}, coexists with a set $K$ of quantum computing clients, collectively training a distributed quantum neural network model (QNN)~\cite{cong2019quantum}. Each client possesses a local dataset comprising quantum states ${|\psi_m\rangle}$ and corresponding labels ${y_m}$ for $m = 1,...,M_k$, which serve as the foundation for training an individual QCNN with learnable parameters $\theta_k$.

The model parameter vector, denoted as $\theta_k$, encapsulates all learnable parameters essential for both quantum circuit operations and classical components. Our principal objective lies in optimizing the QNN model's performance, enabling accurate predictions based on the quantum states within each client's dataset. This innovative approach sets the stage for exploring the intricate intersection of quantum computing and federated learning, opening avenues for transformative advancements in QFL.
 For simplicity and tractability, we utilize the mean squared error (MSE) loss function: 

\begin{equation}
J(\theta_k) = \frac{1}{2M_k}\sum_{m=1}^{M_k} (y_m - f_{\theta_k}(|\psi_m\rangle))^2
\end{equation}
and adopt locally via the VQC~\cite{yano2021efficient}. In the proposed approach, each client device has access to its own dataset and trains its model locally. The primary goal is along the lines of FL design principles as to collectively train a global model without sharing raw data among clients, thereby preserving data privacy \cite{mcmahan2017communication}. 

\begin{figure}[!h]
\centering
\includegraphics[width=0.8\columnwidth]{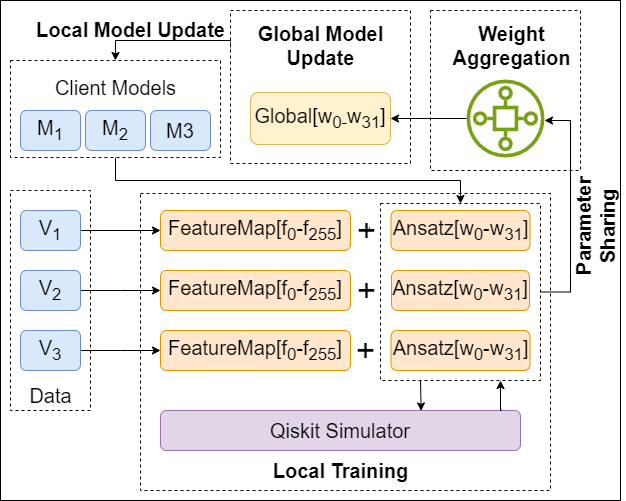}
\caption{\rm \textit{Details of the QFL framework.} Each client possesses unique data (V1, V2, V3), which undergoes transformation into quantum states using a Feature Map of size 255. These quantum states are then processed by a parameterized quantum circuit (Ansatz) with a depth of 3 and 32 weights. Local training using these quantum components is performed using the Qiskit Simulator. After training, the updated weights from the Ansatz are sent to the server for aggregation, where parameters from all clients are combined, and finally, the aggregated global weights are returned to the clients for local model updates.}
\label{fig:qfl_implementation_process}
\end{figure}

Figure \ref{fig:qfl_implementation_process} shows the federation process of the proposed QFL, which involves the following stages:

\noindent \textbf{Local Training}: Each client uses its quantum circuit to process its part of the dataset. The local quantum model is then trained using this data.\\
\textbf{Parameter Sharing}: Upon completion of training in each epoch, clients send their quantum circuit parameters (model weights) to a central server.\\
 \textbf{Aggregation}: The server aggregates these local updates using techniques like simple averaging, weighted averaging, or best pick \cite{mcmahan2017communication}. This aggregation results in an updated global model.\\
 \textbf{Global Model Update}: The updated weights of the global model are sent back to the client devices.\\
 \textbf{Local Model Update}: Each client updates its local model using global weights. The weight updation formula evolves as
\begin{equation}
\mathrm{\theta^{h+1}} = \alpha \times \mathrm{\theta^{h}} + (1 - \alpha) \times \mathrm{global(\theta^{h+1})}
\end{equation}
where $\alpha$ is a weighting factor determined by the current epoch and specific to each client.

\subsection{ Analyzing Aggregation Schemes}

 Three weight aggregation schemes are employed to refine the global model by managing client updates effectively. Firstly, Simple Averaging calculates the average of all local model updates. Mathematically, the new global model parameters $\theta^{h+1}$ are determined as the mean of parameters from individual client models, represented by $\theta_k^h$ for the $k^{th}$ client's model in the $h^{th}$ iteration.

\begin{equation}
\theta^{h+1} = \frac{1}{K} \sum_{k=1}^K \theta_k^h
\end{equation}
This method assumes equal contribution from all clients, regardless of the size or diversity of their local datasets.

\textit{Weighted Averaging} improves the aggregation process by considering the relative importance of each client's update. It assigns weights based on the significance of each client's contribution, resulting in a weighted sum of individual client model updates. This approach allows for a more tailored integration of contributions, accounting for variations in client performance. In this technique, each client's update is weighted by a factor $w_k$ reflecting its significance in the learning process as

\begin{equation}
    \theta^{h+1} = \sum_{k=1}^K w_k \theta_k^h,
\end{equation}
where $\theta^{h+1}$ represents the updated parameters of the model after the $h$-th iteration. The parameters are updated as a weighted sum of the parameters from $K$ client models, denoted by $\theta_k^h$, where each model's contribution is scaled by a weight $w_k$. 

Finally, the \textit{Best Pick} scheme adopts a selective approach, incorporating updates only from clients meeting a predefined performance criterion. The resulting global model is a weighted sum of selected client updates, prioritizing accuracy and reliability. By setting a performance threshold, only updates surpassing this threshold contribute to the global model. The global model update can be represented as:

\begin{equation}
    \theta^{h+1} = \sum_{k \in \mathcal{S}} w_k \theta_k^h
\end{equation}
where summation is taken over the set $\mathcal{S}$, which represents the subset of clients whose updates satisfy the criterion. The weighting factor $w_k$ assigned to each client model's parameters, $\theta_k^h$, is determined by the performance of the client, emphasizing updates from clients deemed more accurate or reliable, potentially enhancing the overall performance and efficiency of the model.



\subsubsection{Datasets} In our experiments, we use a genomic dataset to train a decentralized QFL model, crucial to understanding genomic sequences, consisting of labeled data points $(X, Y)$ representing different genomic characteristics. We chose this dataset due to its sufficient sample size for algorithm testing and the promising potential of genomics for future QFL applications. Our algorithm, demonstrated on the IBM Cloud platform, utilizes multiple QML models and follows the essential steps outlined in Algorithm~\ref{algorithm:qfl_ibm_cloud}, encompassing data preparation, model training, and aggregation of updates from multiple clients. Each client, in our proposed QFL implementation, utilizes a quantum circuit to process genomic data and update model weights based on its local dataset, subsequently aggregated by the server to refine the global model, ensuring data privacy while leveraging collective learning.

\begin{algorithm}[t]
\caption{Proof of Concept QFL over IBM Cloud}

\label{algorithm:qfl_ibm_cloud}

\textbf{Input}: Data $(X, Y)$, Max~ iterations $I$, Initial client weights ${w_{i}}_{i=1}^{n}$, Initial global weights $w_g$, Number of clients $n$\\

\begin{algorithmic}[1]

\State{\textbf{Procedure Initialize}}

\State Split genomic data into user-specific datasets:

\State $(x_{i}, y_{i}) \sim (X, Y) \quad \text{for } i = 1, \ldots, n$

\end{algorithmic}

\noindent -------------------------------------------

\begin{algorithmic}[1]
\State{\textbf{Procedure QFL Learning with Training}}

\For{$epoch = 1, \ldots, I$}

\State Select a subset of participating clients, say $m$ clients

\For{each client $i$ in the selected subset}

\State Initialize client-specific quantum circuit $Q_i$

\State Encode client data $x_{i}$ into quantum states $X_{q_i}$

\State Optimize client's model to minimize loss and update weights: $w_{i}$

\State $w_{i} = \text{minimize loss}(f_{Q_i}(X_{q_i}), y_{i})$

\State Upload local updates $w_{i}$ to the global server

\EndFor

\State Aggregate local updates at the global server:


\State Broadcast updated global weights $w_g$ to all clients

\For{each client $i$}

\State Update local model weights with global weights: 

\EndFor

\EndFor

\end{algorithmic}

\end{algorithm}

\subsection{Encoding driven QFL}

\begin{figure}[t]
\centering
\includegraphics[width=0.8\columnwidth]{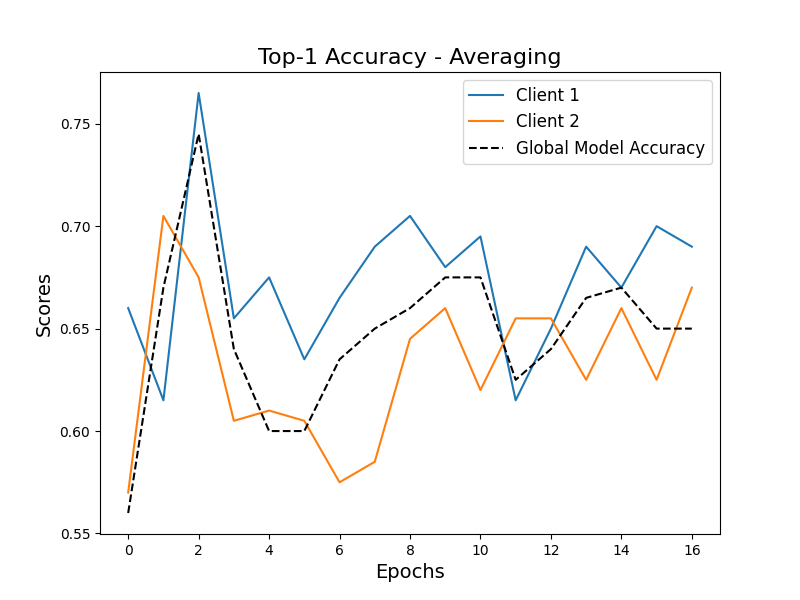}
\caption{\rm Comparision of the evolution of Top-1 Accuracy over epochs for the global model and clients models using the averaging technique.}
\label{fig:averaging_top1_clients_global}
\vspace{-6 mm}
\end{figure}
\begin{figure}[!h]
\centering
\includegraphics[width=0.8\columnwidth]{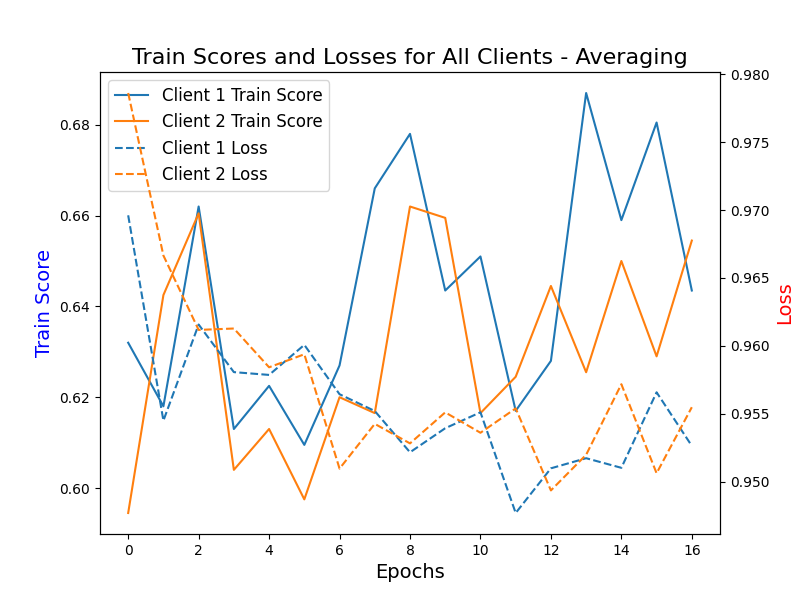}
\caption{\rm Evolution of Top-1 Accuracy and Loss over epochs for  clients using the averaging technique.}
\label{fig:averaging_top1_clients}
\vspace{-8 mm}
\end{figure}


In our implementation of QFL, data encoding stands out as a critical step, transforming classical data into a format compatible with quantum processing. 
Our QFL implementation using a genomic dataset showcases the potential of quantum facilities to manage complex, high-dimensional data. By employing amplitude encoding, we simplify model training with our 200-feature dataset, leveraging only a fraction of the qubits typically required. This approach proves particularly advantageous in genomic data analysis, where traditional computing methods may struggle with data scale and complexity.

We consider three primary encoding techniques: basis encoding, amplitude encoding, and angle encoding, as follows. Basis encoding, while straightforward, suffers from inefficiencies in qubit usage, particularly for larger datasets. Amplitude encoding emerges as a more efficient alternative, especially for handling substantial data volumes. This technique capitalizes on the amplitudes of a quantum state to represent data points, offering a compact representation that can encode vast amounts of information with minimal qubits. For instance, our 200-feature genomic dataset can be effectively encoded using just 8 qubits, making amplitude encoding a practical choice within current quantum computing limitations. This method proves invaluable in quantum machine learning, offering a streamlined approach to represent high-dimensional data.

In contrast, angle encoding, akin to basis encoding in its reliance on one qubit per feature, falls short in scalability for larger datasets due to its demanding qubit requirements. Given these considerations, we opt for amplitude encoding, which efficiently utilizes qubits and aligns with the 25-qubit limit of Qiskit's \texttt{aer\_simulator} utilized in our research.

As shown in Step 6, our Algorithm~\ref{algorithm:qfl_ibm_cloud} efficiently uses quantum circuits for data encoding and processing, presenting a novel methodology for genomic data analysis within a federated learning framework. Utilizing the IBM Cloud platform for quantum computing, our implementation offers scalability and accessibility, potentially paving the way for transformative advancements in genomic research.

\subsection{Performance Evaluation}

\begin{figure}[t]
\centering
\includegraphics[width=0.8\columnwidth]{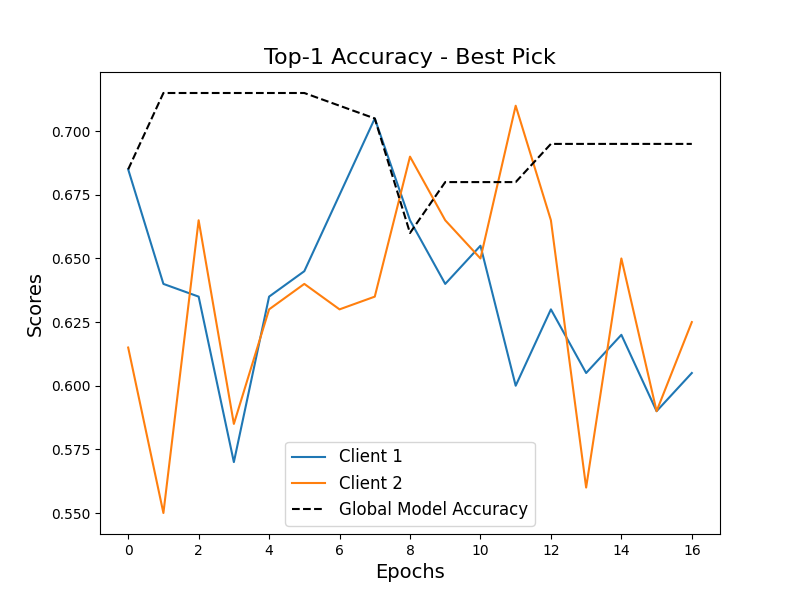}
\caption{\rm Evolution of Top-1 Accuracy for the global model and individual clients using the Best Pick scheme.}
\label{fig:best_pick_top1_clients_global}
\vspace{-5 mm}
\end{figure}

\begin{figure}[t]
\centering
\includegraphics[width=0.8\columnwidth]{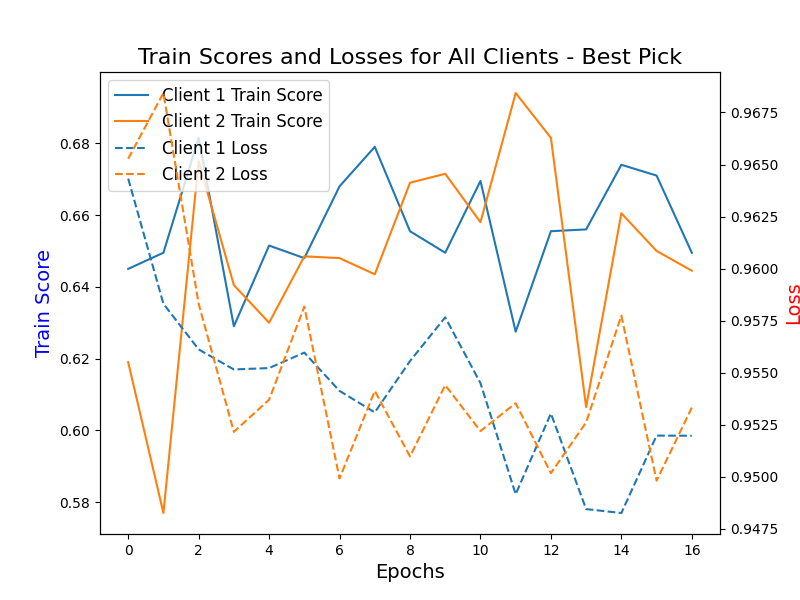}
\caption{\rm Temporal evolution of the training scores and losses for all clients using the Best Pick.}
\label{fig:best_pick_train_loss}
\vspace{-6 mm}
\end{figure}

\begin{figure}[t]
\centering
\includegraphics[width=0.8\columnwidth]{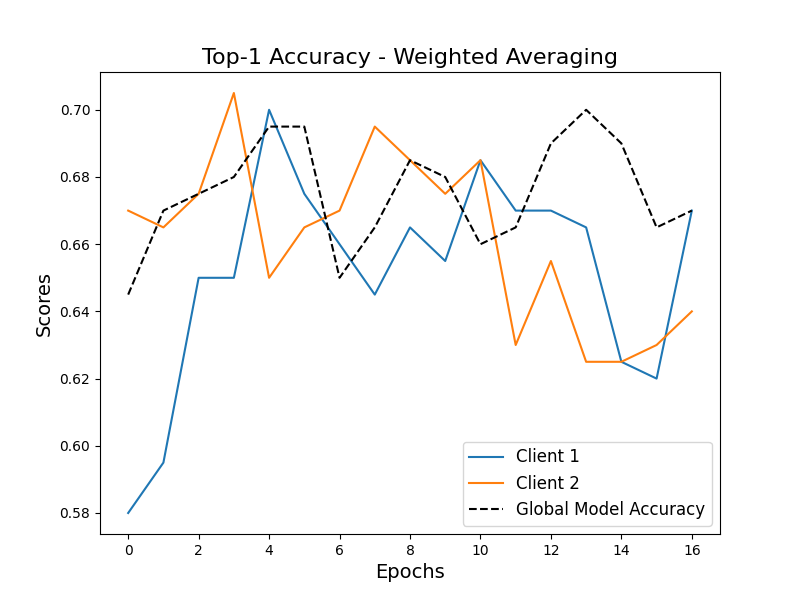}
\caption{\rm Temporal evolution of the Top-1 Accuracy for the global model and individual clients using the Weighted Averaging.}
\label{fig:weighted_averaging_top1_clients_global}
\vspace{-8 mm}
\end{figure}

\begin{figure}[b]
\centering
\includegraphics[width=0.8\columnwidth]{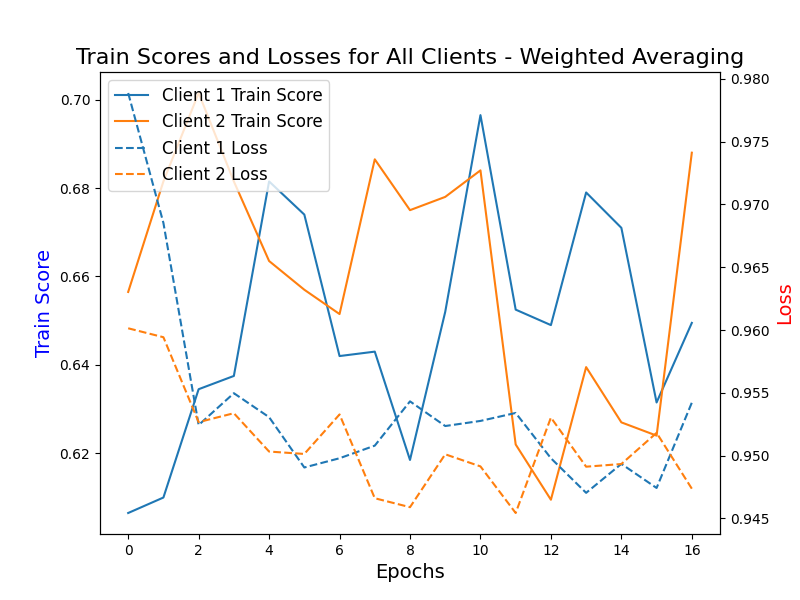}
\caption{\rm Temporal evolution of training scores and losses for all clients using the Weighted Averaging.}
\label{fig:weighted_averaging_train_loss}
\vspace{-10 mm}
\end{figure}

We conducted a series of experiments to train the developed Quantum Federated Learning (QFL) framework using three distinct global aggregation methods: \textit{Averaging, Best Pick, and Weighted Averaging}. The results of these experiments are depicted in Figures \ref{fig:averaging_top1_clients_global} to \ref{fig:weighted_averaging_train_loss} and are detailed below.

In Figure \ref{fig:averaging_top1_clients_global} averaging method exhibited varying Top-1 accuracy between clients, with the global model's accuracy typically positioned between the clients' accuracies. Both the high and low performances of the individual client models directly influenced the global model, placing its accuracy within the range of the client's accuracy. The accuracy and training loss patterns for all clients demonstrated in Figure \ref{fig:averaging_top1_clients} exhibit a consistent decrease in loss and a tractable improvement in accuracy over epochs.

In contrast, the best pick method demonstrated a more selective behavior in Figure \ref{fig:best_pick_top1_clients_global} of our experiments, often resulting in the global model's accuracy surpassing individual clients' accuracies. This observation suggests that the global model performance was driven by the highest-performing clients, with minimal or no contributions from poorly performing ones. The decreasing loss and increasing accuracy observed in Figure~\ref{fig:best_pick_train_loss} over epochs indicated progressive improvements in the training of clients.

With weighted averaging, see the results in Figures \ref{fig:weighted_averaging_top1_clients_global} and \ref{fig:weighted_averaging_train_loss}. The QFL performance closely matched the highest performing clients. This approach minimally affected the global model's performance by poor-performing clients while significantly benefiting from higher-performing ones. This method ensures that the global model takes advantage of the strengths of the higher performing clients while mitigating the impact of the lower performing ones. Training scores and losses exhibited a steady trend of reduced loss and improved accuracy.
\section{Conclusion}
This article highlights the efficacy of different aggregation methods in the operationalization of QFL over Qiskit, demonstrating how data encoding and weighted averaging enhance overall performance by leveraging the strengths of high-performing clients while mitigating the impact of low-performing ones. More research is needed to refine QFL techniques, Qiskit implementation, and transition from simulators to real quantum hardware for successful implementation on cloud computing infrastructure.
\bibliographystyle{ieeetr}
\bibliography{References}


\end{document}